\documentclass[11pt,a4paper]{article}
\usepackage[hyperref]{emnlp2018}
\usepackage{times}
\usepackage{latexsym}
\usepackage{arydshln}
\usepackage{booktabs}
\usepackage{graphicx}
\usepackage{float}
\usepackage{color}
\usepackage{amsmath,amsfonts,amssymb,bbm,epsfig,bm}
\usepackage[ruled,noresetcount,linesnumbered]{algorithm2e}
\usepackage{listings}
\usepackage{subfig}
\usepackage{url}
\usepackage{spverbatim}
\usepackage{multirow}
\usepackage{pbox}
\usepackage{tikz}
\usepackage{pifont}
\usepackage{xspace}
\usepackage{subfig}
\usepackage{physics}
\usepackage[nameinlink,capitalize]{cleveref}
\usepackage{scalerel}
\usepackage[font=small]{caption}
\usepackage{titlesec}
\usepackage{url}
\usepackage{todonotes}
\usepackage{enumitem}

\newcommand{\bb}[1]{\mathbf{#1}}

\aclfinalcopy 

\renewcommand{\tt}[1]{\fontfamily{cmtt}\selectfont #1}

\newcommand\mr{\bm{z}}
\newcommand\x{\bm{x}}

\newcommand\applyconstrn{\textsc{ApplyConstr}}

\newcommand{\ie}{{\emph{i.e.}},\xspace}

\newcommand{\eg}{{\emph{e.g.}},\xspace}
\newcommand{\etc}{\emph{etc.}\xspace}

\def\model/{\textsc{Tranx}}
\def\atis/{\textsc{Atis}}
\def\django/{\textsc{Django}}
\def\wikisql/{\textsc{WikiSQL}}
\def\geo/{\textsc{Geo}}
\def\sq/{{\sc Seq2Tree}}

\titlespacing{\paragraph}{%
  0pt}{
  0.3\baselineskip}{
  1em}

\title{\textbf{\model/}: A Transition-based Neural Abstract Syntax Parser for Semantic Parsing and Code Generation}

\author{Pengcheng Yin, Graham Neubig \\
  Language Technologies Institute \\
  Carnegie Mellon University \\
  {\tt \{pcyin,gneubig\}@cs.cmu.edu}}

\date{}

\begin{document}
\maketitle
\begin{abstract}
 We present \model/, a transition-based neural semantic parser that maps natural language (NL) utterances into formal meaning representations (MRs).
 \model/ uses a transition system based on the \emph{abstract syntax description language} for the target MR, which gives it two major advantages: 
 (1) it is highly accurate, using information from the syntax of the target MR to constrain the output space and model the information flow, and 
 (2) it is highly generalizable, and can easily be applied to new types of MR by just writing a new abstract syntax description corresponding to the allowable structures in the MR.
 Experiments on four different semantic parsing and code generation tasks show that our system is generalizable, extensible, and effective, registering strong results compared to existing neural semantic parsers.\footnote{Available at \href{https://github.com/pcyin/tranX}{\tt https://github.com/pcyin/tranX}. An earilier version is used in~\citet{yin18acl}.}
\end{abstract}

\section{Introduction}


Semantic parsing is the task of transducing natural language (NL) utterances into formal meaning representations (MRs).
The target MRs can be defined according to a wide variety of formalisms.
This include linguistically-motivated semantic representations that are designed to capture the meaning of any sentence such as $\lambda$-calculus~\citep{ZettlemoyerC05} or the abstract meaning representations~\citep{banarescu13amr}.
Alternatively, for more task-driven approaches to semantic parsing, it is common for meaning representations to represent executable programs such as SQL queries~\citep{DBLP:journals/corr/abs-1709-00103}, robotic commands~\cite{artzi-zettlemoyer:2013:TACL}, smart phone instructions~\cite{DBLP:conf/acl/QuirkMG15}, and even general-purpose programming languages like Python~\citep{yin17acl,rabinovich17syntaxnet} and Java~\cite{DBLP:conf/acl/LingBGHKWS16}.

Because of these varying formalisms for MRs, the design of semantic parsers, particularly neural network-based ones has generally focused on a small subset of tasks --- in order to ensure the syntactic well-formedness of generated MRs, a parser is usually specifically designed to reflect the domain-dependent grammar of MRs in the structure of the model~\citep{DBLP:journals/corr/abs-1709-00103,xu2017sqlnet}.
To alleviate this issue, there have been recent efforts in neural semantic parsing with general-purpose grammar models~\citep{DBLP:conf/acl/XiaoDG16,dong18coarsefine}.
\citet{yin17acl} put forward a neural sequence-to-sequence model that generates tree-structured MRs using a series of tree-construction actions, guided by the task-specific context free grammar provided to the model \textit{a priori}.
\citet{rabinovich17syntaxnet} propose the abstract syntax networks (ASNs), where domain-specific MRs are represented by abstract syntax trees (ASTs,~\autoref{fig:ast_gen_example} {\it Left}) specified under the abstract syntax description language (ASDL) framework~\citep{wang97asdl}.
An ASN employs a modular architecture, generating an AST using specifically designed neural networks for each construct in the ASDL grammar.

Inspired by this existing research, we have developed \model/, a \textbf{TRAN}sition-based abstract synta\textbf{X} parser for semantic parsing and code generation.
\model/ is designed with the following principles in mind:

\begin{figure*}[ht]
  \centering
  \includegraphics[width=\textwidth]{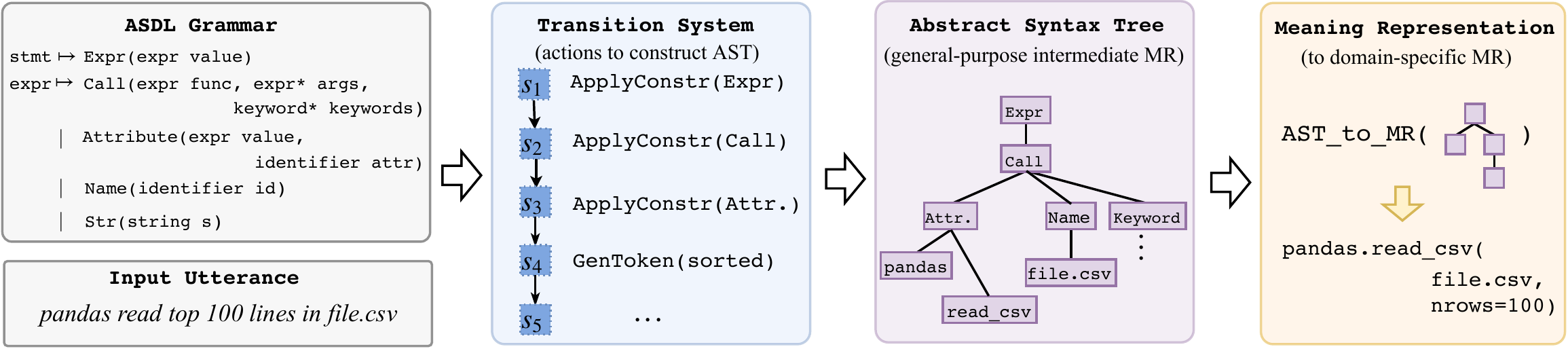}
  \caption{Workflow of \model/}
  \label{fig:system}
\end{figure*}

\begin{itemize}[leftmargin=*]
\setlength\itemsep{0.2em}
  \item \textbf{Generalization ability} \model/ employs ASTs as a general-purpose intermediate meaning representation, and the task-dependent grammar is provided to the system as external knowledge to guide the parsing process, therefore decoupling the semantic parsing procedure with specificities of grammars.
  \item \textbf{Extensibility} \model/ uses a simple transition system to parse NL utterances into tree-structured ASTs. The transition system is designed to be easy to extend, requiring minimal engineering to adapt to tasks that need to handle extra domain-specific information.
  \item \textbf{Effectiveness} We test \model/ on four semantic parsing (\atis/, \geo/) and code generation (\django/, \wikisql/) tasks, and demonstrate that \model/ is capable of generalizing to different domains while registering strong performance, out-performing existing neural network-based approaches on three of the four datasets (\geo/, \atis/, \django/).
\end{itemize}

\section{Methodology}

Given an NL utterance, \model/ parses the utterance into a formal meaning representation, typically represented as $\lambda$-calculus logical forms, domain-specific, or general-purpose programming languages (e.g., Python).
In the following description we use Python code generation as a running example, where a programmer's natural language intents are mapped to Python source code. 
\autoref{fig:system} depicts the workflow of \model/.
We will present more use cases of \model/ in~\autoref{sec:exp}. 

The core of \model/ is a transition system.
Given an input NL utterance $\x$, \model/ employs the transition system to map the utterance $\x$ into an AST $\mr$ using a series of tree-construction actions (\autoref{sec:model:transition}).
\model/ employs ASTs as the intermediate meaning representation to abstract over domain-specific structure of MRs.
This parsing process is guided by the user-defined, domain-specific grammar specified under the ASDL formalism (\autoref{sec:model:asdl}).
Given the generated AST $\mr$, the parser calls the user-defined function, {\tt AST\_to\_MR($\cdot$)}, to convert the intermediate AST into a domain-specific meaning representation $\bm{y}$, completing the parsing process. 
\model/ uses a probabilistic model $p(\mr|\x)$, parameterized by a neural network, to score each hypothesis AST (\autoref{sec:model:network}). 



\subsection{Modeling ASTs using ASDL Grammar}
\label{sec:model:asdl}

\begin{figure*}[t]
\begin{minipage}[t]{1. \columnwidth}
  \centering
  \vspace{0pt}
  \setlength{\tabcolsep}{0.em} 
  \includegraphics[width=\columnwidth]{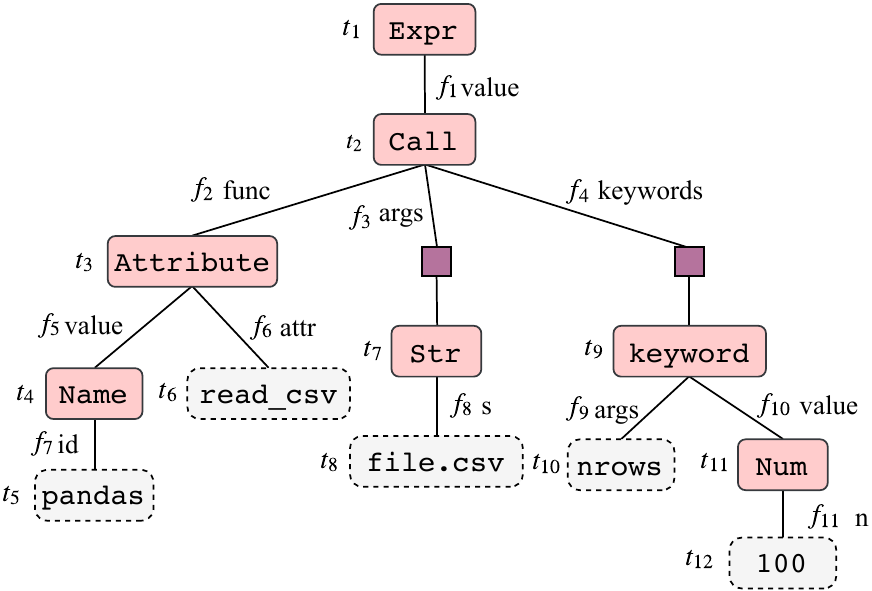}
\end{minipage}%
\hspace{10mm}%
\begin{minipage}[t]{0.9 \columnwidth}
  {\renewcommand{\arraystretch}{0.9} 
  \vspace{0pt}
  \resizebox{\columnwidth}{!}{%
  \setlength\tabcolsep{3pt}
  \begin{tabular}{lll}
  \hline

  \hline
  $\bm{t}$ & $\bm{n_{f_t}}$ & {\bf Action} \\
  \hdashline
  $t_1$ & {\tt root}     & {\tt Expr(expr value)}  \\
  $t_2$ & $f_1$    & {\tt Call(expr func, expr* args, }  \\
      &           & \hfill {\tt keyword* keywords)} \\
  $t_3$ & $f_2$    & {\tt Attribute(expr value, identifier attr)} \\
  $t_4$ & $f_5$  & {\tt Name(identifier id)}  \\
  $t_5$ & $f_7$  & {\sc GenToken}$[\textrm{pandas}]$  \\
  $t_6$ & $f_6$  & {\sc GenToken}$[\textrm{read\_csv}]$  \\
  $t_7$ & $f_3$  & {\tt Str(string s)}  \\ 
  $t_8$ & $f_8$ & {\sc GenToken}$[\textrm{file.csv}]$  \\
  $t_9$ & $f_8$ & {\sc GenToken}$[$\textless/f\textgreater$]$  \\
  $t_{10}$ & $f_3$ & {\sc Reduce} (close the frontier field $f_3$)  \\
  $t_{11}$ & $f_4$ & {\tt keyword(identifier arg, expr value)} \\
  $t_{12}$ & $f_9$ & {\sc GenToken}$[\textrm{nrows}]$ \\
  $t_{13}$ & $f_{10}$ & {\tt Num(object n)} \\
  $t_{14}$ & $f_{11}$ & {\sc GenToken}$[\textrm{1000}]$ \\
  $t_{15}$ & $f_4$ & {\sc Reduce} (close the frontier field $f_4$) \\
  \hline

  \hline
  \end{tabular}}}
\end{minipage}
\caption{\textbf{\textit{Left}} The ASDL AST for the target Python code in~\autoref{fig:system}. Field names are labeled on upper arcs, and indexed as $f_i$. Purple squares denote fields with {\it sequential} cardinality. Grey nodes denote primitive identifier fields. Fields are labeled with time steps at which they are generated. \textbf{\textit{Right}} The action sequence used to construct the AST. Each action is labeled with its frontier field $n_{f_t}$. {\sc ApplyConstr} actions are represented by their constructors.}
\vspace{-3mm}
\label{fig:ast_gen_example}
\end{figure*}

\model/ uses ASTs as the general-purpose, intermediate semantic representation for MRs.
ASTs are commonly used to represent programming languages, and can also be used to represent other tree-structured MRs (\eg $\lambda$-calculus).
The ASDL framework is a grammatical formalism to define ASTs.
See~\autoref{fig:system} for an excerpt of the Python ASDL grammar.
\model/ provides APIs to read such a grammar from human-readable text files.

An ASDL grammar has two basic constructs: \emph{types} and \emph{constructors}.
A \emph{composite} type is defined by the set of constructors under that type.
For example, the {\tt stmt} and {\tt expr} composite types in~\autoref{fig:system} refer to Python statements and expressions, repectively, each defined by a series of constructors.
A constructor specifies a language construct of a particular type using its \emph{fields}.
For instance, the {\tt Call} constructor under the composite type {\tt expr} denotes function call expressions, and has three fields: {\tt func}, {\tt args} and {\tt keywords}.
Each field in a constructor is also strongly typed, which specifies the type of value the field can hold.
A field with a composite type can be instantiated by constructors of the same type. For example, the {\tt func} field above can hold a constructor of type {\tt expr}.
There are also fields with \emph{primitive} types, which store values.
For example, the {\tt id} field of {\tt Name} constructor has a primitive type {\tt identifier}, and is used to store identifier names.
And the field {\tt s} in the {\tt Str} (string) constructor hold string literals.
Finally, each field has a cardinality (single, optional $?$ and sequential $*$), denoting the number of values the field holds.

An AST is then composed of multiple constructors, where each node on the tree corresponds to a typed field in a constructor (except for the root node, which denotes the root constructor).
Depending on the cardinality of the field, a node can hold one or multiple constructors as its values.
For instance, the {\tt func} field with single cardinality in the ASDL grammar in~\autoref{fig:system} is instantiated with one {\tt Name} constructor, while the {\tt args} field with sequential cardinality have multiple child constructors.

\subsection{Transition System}
\label{sec:model:transition}

Inspired by~\citet{yin17acl} (hereafter YN17), we develop a transition system that decomposes the generation procedure of an AST into a sequence of tree-constructing \emph{actions}.
We now explain the transition system using our running example. 
\autoref{fig:ast_gen_example} {\it Right} lists the sequence of actions used to construct the example AST.
In high level, the generation process starts from an initial derivation AST with a single root node, and proceeds according to a top-down, left-to-right order traversal of the AST.
At each time step, one of the following three types of actions is evoked to expand the opening \emph{frontier field} $n_{f_t}$ of the derivation:

\textbf{\textsc{ApplyConstr}$[c]$} actions apply a constructor $c$ to the opening composite frontier field which has the same type as $c$, populating the opening node using the fields in $c$.
If the frontier field has sequential cardinality, the action appends the constructor to the list of constructors held by the field.

\textbf{\textsc{Reduce}} actions mark the completion of the generation of child values for a field with optional (?) or multiple ($*$) cardinalities.

\textbf{\textsc{GenToken}}$[v]$ actions populate a (empty) primitive frontier field with a token $v$.
For example, the field $f_7$ on~\autoref{fig:ast_gen_example} has type {\tt identifier}, and is instantiated using a single \textsc{GenToken} action.
For fields of {\tt string} type, like $f_8$, whose value could consists of multiple tokens (only one shown here), it can be filled using a sequence of \textsc{GenToken} actions, with a special $\textsc{GenToken}[${\tt </f>}$]$ action to terminate the generation of token values.

The generation completes once there is no frontier field on the derivation.
\model/ then calls the user specified function {\tt AST\_to\_MR($\cdot$)} to convert the generated intermediate AST $\mr$ into the target domain-specific MR $\bm{y}$.
\model/ provides various helper functions to ease the process of writing conversion functions.
For example, our example conversion function to transform ASTs into Python source code contains only 32 lines of code.
\model/ also ships with several built-in conversion functions to handle MRs commonly used in semantic parsing and code generation, like $\lambda$-calculus logical forms and SQL queries.

\vspace{-1.5mm}
\subsection{Computing Action Probabilities $p(\mr|\x)$}
\vspace{-1.5mm}
\label{sec:model:network}
Given the transition system, the probability of an $\mr$ is decomposed into the probabilities of the sequence of actions used to generate $\mr$
\begin{equation*}\setlength\abovedisplayskip{5pt}\setlength\belowdisplayskip{5pt}
  p(\mr|\x) = \prod_{t} p(a_t | a_{<t}, \x),
\end{equation*}
Following YN17, we parameterize the transition-based parser $p(\mr|\x)$ using a neural encoder-decoder network with augmented recurrent connections to reflect the topology of ASTs.

\paragraph{Encoder} The encoder is a standard bidirectional Long Short-term Memory (LSTM) network, which encodes the input utterance $\x$ of $n$ tokens, $\{ x_i \}_{i=1}^n$ into vectorial representations $\{ \mathbf{h} \}_{i=1}^n$.

\paragraph{Decoder} The decoder is also an LSTM network, with its hidden state $\mathbf{s}_t$ at each time temp given by
\begin{equation*}\setlength\abovedisplayskip{5pt}\setlength\belowdisplayskip{5pt}
  \bb{s}_t = f_{\textrm{LSTM}}([\bb{a}_{t-1}: \bb{\tilde{s}}_{t-1}: \bb{p}_t], \bb{s}_{t-1}),
\end{equation*}
where $f_{\textrm{LSTM}}$ is the LSTM transition function, and $[:]$ denotes vector concatenation.
$\bb{a}_{t-1}$ is the embedding of the previous action. We maintain an embedding vector for each action. 
$\bb{\tilde{s}}_{t}$ is the attentional vector defined as in~\citet{luong2015effective}
\begin{equation*}\setlength\abovedisplayskip{5pt}\setlength\belowdisplayskip{5pt}
  \bb{\tilde{s}}_t = \tanh(\bb{W}_c[\bb{c}_t: \bb{s}_t]).
  \label{eq:att_vec}
\end{equation*}
where $\bb{c}_t$ is the context vector retrieved from input encodings $\{ \bb{h}_i \}_{i=1}^n$ using attention.

\paragraph{Parent Feeding} $\bb{p}_t$ is a vector that encodes the information of the parent frontier field $n_{f_t}$ on the derivation, which is a concatenation of two vectors: the embedding of the frontier field $\bb{n}_{f_t}$, and $\bb{s}_{p_t}$, the decoder's state at which the constructor of $n_{f_t}$ is generated by the \applyconstrn~action. Parent feeding reflects the topology of tree-structured ASTs, and gives better performance on generating complex MRs like Python code (\autoref{sec:exp}).

\begin{figure}[t]
  \centering
\begin{lstlisting}[basicstyle=\fontfamily{cmtt}\small,columns=fullflexible,frame=bt]
expr 
 = Variable(var variable)
 | Entity(ent entity)
 | Number(num number)
 | Apply(pred predicate, expr* arguments)
 | Argmax(var variable, expr domain, expr body)
 | Argmin(var variable, expr domain, expr body)
 | Count(var variable, expr body)
 | Exists(var variable, expr body)
 | Lambda(var variable, var_type type, expr body)
 | Max(var variable, expr body)
 | Min(var variable, expr body)
 | Sum(var variable, expr domain, expr body)
 | The(var variable, expr body)
 | Not(expr argument)
 | And(expr* arguments)
 | Or(expr* arguments)
 | Compare(cmp_op op, expr left, expr right)

cmp_op = Equal | LessThan | GreaterThan
\end{lstlisting}
  \vspace{-2mm}
  \caption{The $\lambda$-calculus ASDL grammar for \geo/ and \atis/, defined in~\citet{rabinovich17syntaxnet}}
  \label{fig:asdl_atis}
  \vspace{-5mm}
\end{figure}

\paragraph{Action Probabilities} 
The probability of an~\applyconstrn$[c]$~action with embedding $\bb{a}_c$ is\footnote{\textsc{Reduce} is treated as a special \applyconstrn\ action.}
\begingroup
\setlength\abovedisplayskip{5pt}
\setlength\belowdisplayskip{5pt}
\begin{multline}
  p(a_t  = \textsc{ApplyConstr}[c]|a_{<t}, \x) \\
  = \textrm{softmax}(\bb{a}_c^\intercal \bb{W} \bb{\tilde{s}}_t)
  \label{eq:actionprob:applyconstr}
\end{multline}
\endgroup
For \textsc{GenToken} actions, we employ a hybrid approach of generation and copying, allowing for out-of-vocabulary variable names and literals (\eg ``\textit{file.csv}'' in~\autoref{fig:system}) in $\x$ to be directly copied to the derivation. Specifically, the action probability is defined to be the marginal probability
\begingroup
\setlength\abovedisplayskip{5pt}
\setlength\belowdisplayskip{5pt}
\begin{multline*}
  p(a_t = \textsc{GenToken}[v]|a_{<t}, \x) \\
   = p(\textrm{gen}| a_{t}, \x) p(v|\textrm{gen}, a_{t}, \x) + \\
     p(\textrm{copy}| a_{t}, \x) p(v|\textrm{copy}, a_{t}, \x)
\end{multline*}
\endgroup
The binary probability $p(\textrm{gen}|\cdot)$ and $p(\textrm{copy}|\cdot)$ is given by $\textrm{softmax}(\bb{W}\bb{\tilde{s}}_t)$.
The probability of generating $v$ from a closed-set vocabulary, $p(v|\textrm{gen}, \cdot)$ is defined similarly as~\cref{eq:actionprob:applyconstr}.
The copy probability of copying the $i$-th word in $\x$ is defined using a pointer network~\citep{DBLP:conf/nips/VinyalsFJ15}
\begin{equation*}\setlength\abovedisplayskip{5pt}\setlength\belowdisplayskip{5pt}
  p(x_i|\textrm{copy}, a_{<t}, \x) = \textrm{softmax} ( \bb{h}_i^\intercal \bb{W} \bb{\tilde{s}}_t ).
\end{equation*}

\section{Experiments}
\label{sec:exp}

\vspace{-1mm}
\subsection{Datasets}
\vspace{-1mm}
To demonstrate the generalization and extensibility of \model/, we deploy our parser on four semantic parsing and code generation tasks.

\vspace{-1mm}
\subsubsection{Semantic Parsing}
\vspace{-1mm}

We evaluate on \geo/ and \atis/ datasets.
\geo/ is a collection of 880 U.S.~geographical questions (\eg \textit{``Which states border Texas?''}), and \atis/ is a set of 5,410 inquiries of flight information (\eg \textit{``Show me flights from Dallas to Baltimore''}). The MRs in the two datasets are defined in $\lambda$-calculus logical forms (\eg~``{\tt lambda $x$ (and (state $x$) (next\_to $x$ texas))}'' and ``{\tt lambda $x$ (and (flight $x$ dallas) (to $x$ baltimore))}'').
We use the pre-processed datasets released by~\citet{DBLP:conf/acl/DongL16}.
We use the ASDL grammar defined in~\citet{rabinovich17syntaxnet}, as listed in~\autoref{fig:asdl_atis}.

\begin{figure}[t]
  \centering
\begin{lstlisting}[basicstyle=\fontfamily{cmtt}\small,columns=fullflexible,frame=bt]
stmt = Select(agg_op? agg, idx column_idx, 
                            cond_expr* conditions)
cond_expr = Condition(cmp_op op, idx column_idx, 
                                      string value)
agg_op = Max | Min | Count | Sum | Avg
cmp_op = Equal | GreaterThan | LessThan | Other
\end{lstlisting}
  \vspace{-2mm}
  \caption{The ASDL grammar for \wikisql/}
  \label{fig:asdl_wikisql}
  \vspace{-3mm}
\end{figure}

\vspace{-1mm}
\subsubsection{Code Generation}
\vspace{-1mm}

We evaluate \model/ on both general-purpose (Python, \django/) and domain-specific (SQL, \wikisql/) code generation tasks.
The \django/ dataset~\citep{DBLP:conf/kbse/OdaFNHSTN15} consists of 18,805 lines of Python source code extracted from the Django Web framework, with each line paired with an NL description. 
Code in this dataset covers various real-world use cases of Python, like string manipulation, I/O operation, exception handling, \etc

\wikisql/~\citep{DBLP:journals/corr/abs-1709-00103} is a code generation task for \emph{domain-specific} languages (\ie SQL). It consists of 80,654 examples of NL questions (\eg \textit{``What position did Calvin Mccarty play?''}) and annotated SQL queries (\eg ``{\tt SELECT Position FROM Table WHERE Player = Calvin Mccarty}''). Different from other datasets, each example also has a table extracted from Wikipedia, and the SQL query is executed against the table to get an answer.

\paragraph{Extending \model/ for \wikisql/} 
In order to achieve strong results, existing parsers, like most models in~\autoref{tab:exp:resuts:wiki}, use specifically designed architectures to reflect the syntactic structure of SQL queries.
We show that the transition system used by \model/ can be easily extended for \wikisql/ with minimal engineering, while registering strong performance.
First, we use define a simple ASDL grammar following the syntax of SQL (\autoref{fig:asdl_wikisql}).
We then augment the transition system with a special \textsc{GenToken} action, $\textsc{SelColumn}[k]$. 
A $\textsc{SelColumn}[k]$ action is used to populate a primitive {\tt column\_idx} field in {\tt Select} and {\tt Condition} constructors in the grammar by selecting the $k$-th column in the table.
To compute the probability of $\textsc{SelColumn}[k]$ actions, we use a pointer network over column encodings, where the column encodings are given by a bidirectional LSTM network over column names in an input table. 
This can be simply implemented by overriding the base {\tt Parser} class in \model/ and modifying the functions that compute action probabilities.

\begin{table}[tb]
\small
  \centering
  \begin{tabular}{lcc}
  \hline

  \hline
  \textbf{Methods} & \textbf{\geo/} & \textbf{\atis/} \\
  \hline
    ZH15 \citep{zhao15type} & 88.9 & 84.2 \\
    ZC07 \citep{DBLP:dblp_conf/emnlp/ZettlemoyerC07} & 89.0 & 84.6 \\
    WKZ14 \citep{wang-kwiatkowski-zettlemoyer:2014:EMNLP2014} & {\bf 90.4} & {\bf 91.3} \\ \hline

    \hline
    \textbf{Neural Network-based Models} \\
    \sq/ \citep{DBLP:conf/acl/DongL16} & 87.1 & 84.6 \\
    ASN \citep{rabinovich17syntaxnet}  & 85.7 & 85.3 \\
    \quad\quad + supervised attention  &  87.1 & 85.9 \\
  \hline 
    \model/ (w/o parent feeding) & 88.2 & 86.2 \\
    \model/ (w/ parent feeding) & 87.7 & 86.2 \\

  \hline

  \hline
  \end{tabular}
  \caption{Semantic parsing accuracies on \geo/ and \atis/}
  \label{tab:exp:resuts:sp}
  \vspace{-2mm}
\end{table}

\begin{table}[tb]
\small
  \centering
  \begin{tabular}{lc}
  \hline

  \hline
  \textbf{Methods} & \textsc{Acc.}  \\
  \hline
    Phrasal Statistical MT~\citep{DBLP:conf/acl/LingBGHKWS16} & 31.5 \\
    \sq/~\citep{DBLP:conf/acl/DongL16} &  39.4 \\
    \textsc{nmt}~\citep{neubig15lamtram} & 45.1 \\
    \textsc{lpn}~\citep{DBLP:conf/acl/LingBGHKWS16} & 62.3 \\
    YN17~\citep{yin17acl} & 71.6 \\ \hline
    \model/ (w/o parent feeding) & 72.7 \\
    \model/ (w parent feeding)   & \textbf{73.7} \\
  \hline

  \hline
  \end{tabular}
  \caption{Code generation accuracies on \django/}
  \label{tab:exp:resuts:django}
  \vspace{-4mm}
\end{table}

\begin{table}[t]
\small
  \centering
  \begin{tabular}{lcc}
  \hline

  \hline
  \textbf{Methods} & \textsc{Acc}$_\textit{EM}$ & \textsc{Acc}$_\textit{EX}$ \\
  \hline
    Seq2Seq \citep{DBLP:journals/corr/abs-1709-00103} & 23.4 & 35.9 \\
    \sq/ \citep{DBLP:conf/acl/DongL16} & 23.4 & 35.9 \\
    Seq2SQL \citep{DBLP:journals/corr/abs-1709-00103} & 48.3 & 59.4 \\
    SQLNet \citep{xu2017sqlnet} & -- & 68.0 \\
    PT-MAML \citep{huang2018sqlmeta} & 62.8 & 68.0 \\
    TypeSQL \citep{yu18typesql} & -- & {\bf 73.5} \\
  \hline
    \model/ \\
    \quad w/ parent feeding & 62.6 & 71.6 \\
    \quad w/o parent feeding & {\bf 62.9} & 71.7 \\
  \hline

  \hline
    PointSQL \citep{wang17pointsql}$^\dagger$ & 61.5 & 66.8 \\
    TypeSQL+TC \citep{yu18typesql}$^\dagger$ & -- & {\bf 82.6} \\
    STAMP \citep{sun18sql}$^\dagger$ & 60.7 & 74.4 \\
    STAMP+RL \citep{sun18sql}$^\dagger$ & 61.0 & 74.6 \\
  \hline
    \model/ \\
    \quad w par. feed. + answer pruning$^\dagger$ & 68.4 & 78.6 \\
    \quad w/o par. feed. + answer pruning$^\dagger$  & {\bf 68.6} & 78.6 \\
  \hline

  \hline
  \end{tabular}
  \caption{Exact match (EM) and execution (EX) accuracies on \wikisql/. $^\dagger$Methods that use the contents of input tables. }
  \label{tab:exp:resuts:wiki}
  \vspace{-4mm}
\end{table}

\vspace{-1mm}
\subsection{Results}
\vspace{-1mm}

In this section we discuss our experimental results.
All results are averaged over three runs with different random seeds.

\paragraph{Semantic Parsing} \autoref{tab:exp:resuts:sp} lists the results for semantic parsing tasks.
We test \model/ with two configurations, with or without parent feeding~(\autoref{sec:model:network}). Our system outperforms existing neural network-based approaches.
This demonstrates the effectiveness of \model/ in closed-domain semantic parsing.
Interestingly, we found the model without parent feeding achieves slightly better accuracy on \geo/, probably because that its relative simple grammar does not require extra handling of parent information.

\paragraph{Code Generation}
\autoref{tab:exp:resuts:django} lists the results on \django/.
\model/ achieves state-of-the-art results on \django/.
We also find parent feeding yields +1 point gain in accuracy, suggesting the importance of modeling parental connections in ASTs with complex domain grammars (\eg Python).

\autoref{tab:exp:resuts:wiki} shows the results on \wikisql/. We first discuss our standard model which only uses information of column names and do not use the contents of input tables during inference, as listed in the top two blocks in~\autoref{tab:exp:resuts:wiki}.
We find \model/, although just with simple extensions to adapt to this dataset,
achieves impressive results and outperforms many \emph{task-specific} methods.
This demonstrates that \model/ is easy to extend to incorporate task-specific information, while maintaining its effectiveness.
We also extend \model/ with a very simple \emph{answer pruning} strategy, where we execute the candidate SQL queries in the beam against the input table, and prune those that yield empty execution results. 
Results are listed in the bottom two-blocks in~\autoref{tab:exp:resuts:wiki}, where we compare with systems that also use the contents of tables.
Surprisingly, this (frustratingly) simple extension yields significant improvements, outperforming many task-specific models that use specifically designed,
heavily-engineered 
neural networks to incorporate information of table contents.
\vspace{-1.5mm}
\section{Conclusion}
\vspace{-1.5mm}
We present \model/, a transition-based abstract syntax parser. 
\model/ is generalizable, extensible and effective, achieving strong results on semantic parsing and code generation tasks.

\vspace{-1mm}
\section*{Acknowledgements}
\vspace{-1mm}


This material is based upon work supported by the National Science Foundation under Grant No. 1815287. PY would like to thank Junxian He and Li Dong for helpful discussions.


\bibliography{parser}
\bibliographystyle{acl_natbib_nourl}

\end{document}